%% file: main.tex
\documentclass{bmvc2k}
\usepackage[latin9]{inputenc}
\usepackage{xcolor}
\usepackage{array}
\usepackage{float}
\usepackage{multirow}
\usepackage{dsfont}
\usepackage{amsmath}
\usepackage{graphicx}
\usepackage[]
 {hyperref}

\makeatletter

\providecommand{\tabularnewline}{\\}

\usepackage{amssymb}
\usepackage{dsfont}
\usepackage{pifont}
\usepackage{capt-of}
\usepackage{caption}
\title{Unified Framework with Consistency across Modalities \\ for Human Activity Recognition}

\addauthor{Tuyen Tran}{t.tran@deakin.edu.au}{1}
\addauthor{Thao Minh Le}{thao.le@deakin.edu.au}{1}
\addauthor{Hung Tran}{tduy@deakin.edu.au}{1}
\addauthor{Truyen Tran}{truyen.tran@deakin.edu.au}{1}

\addinstitution{
 Applied Artificial Intelligence Institute\\
 Deakin University\\
 Australia
}

\runninghead{Tuyen Tran, Thao Minh Le, Hung Tran, Truyen Tran}{COMPUTER}

\ifdefined\showcaptionsetup
 \PassOptionsToPackage{caption=false}{subfig}
\fi
\usepackage{subfig}
\makeatother

\begin{document}
\title{Unified Compositional Query Machine with Multimodal Consistency for
Video-based Human Activity Recognition}

\maketitle
\maketitle

\thispagestyle{empty}

\global\long\def\ModelName{\text{COMPUTER}}%

\global\long\def\BlockName{\text{HUB}}%

\begin{abstract}
\input{abs.tex}
\end{abstract}

\section{Introduction}

\input{intro.tex}

\section{Related Work}

\input{related.tex}

\section{Method\protect\label{sec:Method}}

\input{method.tex}

\section{Experiments\protect\label{sec:Experiments}}

\input{exp.tex}

\section{Conclusion}

\input{conclusion.tex}

\bibliography{egbib}
 
\end{document}

%% file: abs.tex
Recognizing human activities in videos is challenging due to the spatio-temporal
complexity and context-dependence of human interactions. Prior studies
often rely on single input modalities, such as RGB or skeletal data,
limiting their ability to exploit the complementary advantages across
modalities. Recent studies focus on combining these two modalities
using simple feature fusion techniques. However, due to the inherent
disparities in representation between these input modalities, designing
a unified neural network architecture to effectively leverage their
complementary information remains a significant challenge. To address
this, we propose a comprehensive multimodal framework for robust video-based
human activity recognition. Our key contribution is the introduction
of a novel \emph{compositional query machine,} called $\ModelName$
(\textbf{COMP}ositional h\textbf{U}man-cen\textbf{T}ric qu\textbf{ER}y
machine), a generic neural architecture that models the interactions
between a human of interest and its surroundings in both space and
time. Thanks to its versatile design, $\ModelName$ can be leveraged
to distill distinctive representations for various input modalities.
Additionally, we introduce a consistency loss that enforces agreement
in prediction between modalities, exploiting the complementary information
from multimodal inputs for robust human movement recognition. Through
extensive experiments on action localization and group activity recognition
tasks, our approach demonstrates superior performance when compared
with state-of-the-art methods. Our code is available at: \href{https://github.com/tranxuantuyen/COMPUTER}{https://github.com/tranxuantuyen/COMPUTER}.

%% file: intro.tex
\begin{figure}
\begin{centering}
\includegraphics[width=0.8\columnwidth]{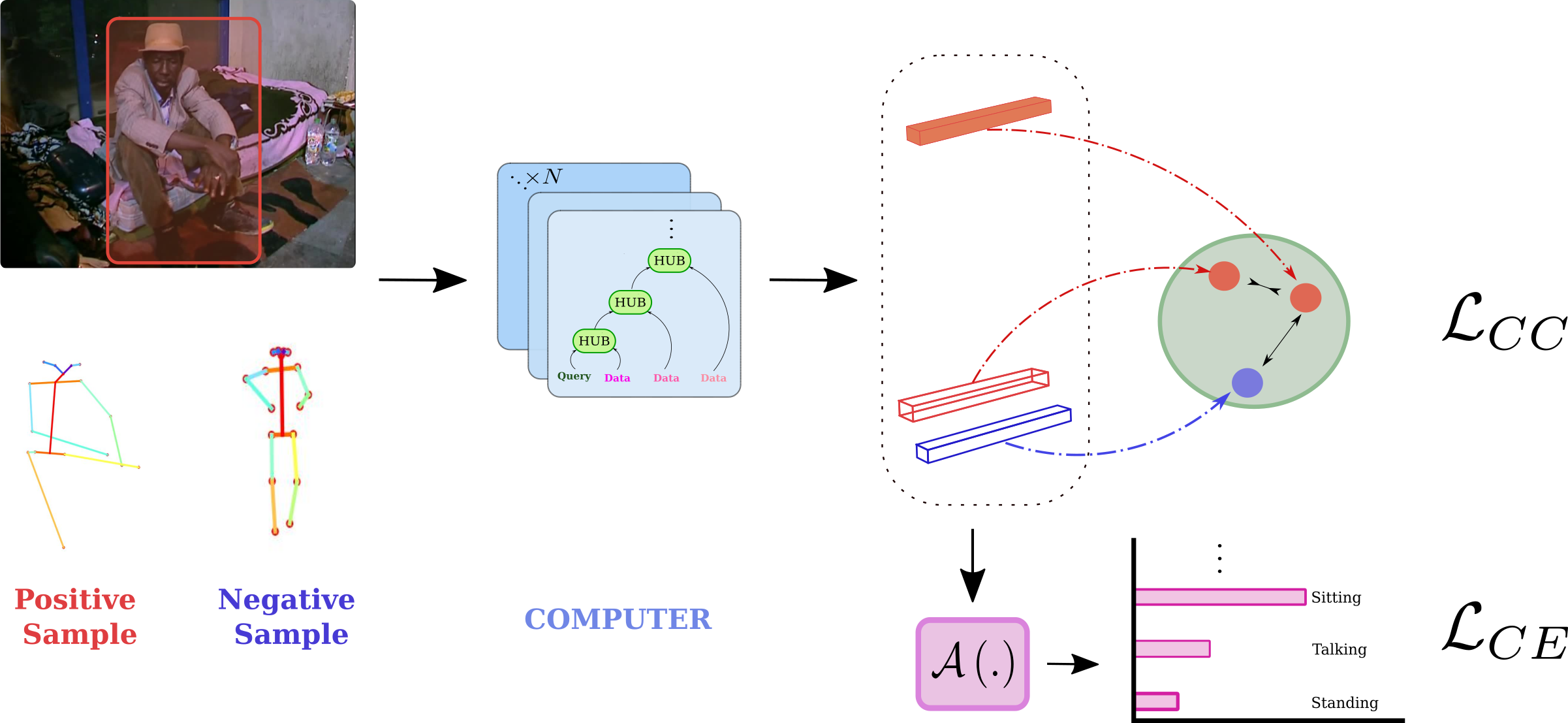}\vspace{4mm}
\par\end{centering}
\caption{\textbf{Method overview:} We use a unified network architecture $\protect\ModelName$
to extract high-level representations of human activity from multi-modal
inputs, including RGB sequences and body key points. The entire frame
is trained end-to-end using a combination of two loss functions: cross-entropy
loss for label prediction and contrastive loss for consistency between
modalities. Notably, our consistency loss maximizes the mutual information
between different input modalities for the same activity in an unsupervised
manner. \protect\label{fig:teaser}}
\end{figure}

Human activity recognition in videos is a crucial area of focus within
the field of Artificial Intelligence (AI), enabling numerous practical
applications in real-world scenarios \cite{sun2018actor,wu2019learning}.
However, this is a challenging task due to the spatio-temporal complexity
and context-dependence of human interactions. These factors require
AI systems to robustly interpret and generalize across a wide range
of behaviors and environmental conditions.

Previous studies have explored representing human activities in consideration
of contextual factors \cite{pan2021actor,wu2020context,sun2018actor}.
Tang \emph{et al.}\cite{tang2020asynchronous} analyze human actions
from an object-centric perspective, modeling the relationships between
humans and the surrounding objects. These works all model the dynamics
of visual scenes by focusing on how the relationships between entities
evolve over time. However, they rely solely on computationally expensive
RGB image sequences, making temporal representation from video data
challenging \cite{duan2022revisiting}. Additionally, using only RGB
data limits the capability to capture subtle body movements. 

Human body key points and skeleton data offer advantages in computational
costs and temporal modeling due to their compactness and robustness
against lighting conditions and scene variations \cite{shi2019skeleton,duan2022revisiting}.
However, skeleton data lacks contextual information, limiting its
capability to represent spatial relationships involved in human-object
interactions.

Given the complementary attributes of RGB and skeleton data, a natural
question arises: \emph{how to design a unified model to effectively
combine these modalities for the task of human activity recognition}?
However, it is not straightforward to build a joint representation
that leverages both modalities' strengths due to their inherent disparities.
HIT \cite{faure2023holistic} was among the earliest attempt to address
this challenge. Their approach involved designing separate components
to process each modality independently, followed by a late fusion
technique. Due to significant structural differences between modalities,
late fusion performed poorly. This is because one modality may negatively
impact the other, ultimately reducing the representational capabilities
of the joint features.

To address the limitations of current methods, we first propose a
unified feature representation framework for multiple modalities in
human activity recognition. Second, we introduce a novel self-supervised
mechanism to ensure consistency in prediction using different modalities
to avoid negative cross-modality impacts within their joint representation.
Overview of the proposed approach is illustrated in Fig.~\ref{fig:teaser}.
To the best of our knowledge, we are the first to propose a generic
and modality-agnostic architecture, along with a novel mechanism for
multi-modal consistency for human activity recognition. To evaluate
the effectiveness of the proposed approach, we conduct intensive experiments
on two human activity recognition tasks: Spatio-Temporal Action Localization
and Group Activity Recognition.

In summary, our contribution is three-fold: (1) Introduction of a
unified compositional query machine for simultaneously handling multi-modal
inputs for the task of human-centric video understanding; (2) Introduction
of a novel mechanism to encourage consistency in prediction across
modalities in a self-supervised manner; (3) Conducting extensive experiments
and analyses across two tasks in human activity recognition in videos.

%% file: related.tex
\subsection{Multi-modal human action recognition }

Prior works on human activity recognition mostly rely solely on RGB
features, discarding valuable information from other modalities. For
instance, skeleton data offers distinct advantages in recognizing
actions that require superior temporal modeling such as running or
driving a car \cite{shi2019skeleton}. Recognizing the benefits of
multi-modal input, some studies have attempted to incorporate additional
modalities beyond RGB features \cite{faure2023holistic,Su2019Improving,sun2018actor,gu2018ava}.
PCSC \cite{Su2019Improving} proposes to use optical flow to capture
motion, designing an inception-like model with an early fusion mechanism
to combine RGB with flow features. In contrast, \cite{sun2018actor}
extracts RGB and motion features using I3D \cite{carreira2017quo},
and then combines them with a late fusion mechanism. Most recently,
HIT \cite{faure2023holistic} utilizes both skeleton data and RGB
features for spatio-temporal action localization using a simple late
fusion technique. While these approaches have shown some benefits
of using multi-modal inputs, neither early fusion or late fusion are
capable of building a joint representation that captures the complementary
advantages across modalities. Different from these works, our approach
uses a novel mechanism to leverage the consistency in prediction across
different modalities for robust human activity recognition. More importantly,
our newly introduced consistency loss allows us to train our proposed
method in an unsupervised manner without the need for additional training
data.

\subsection{Contrastive self-supervised learning}

Contrastive self-supervised learning has gained popularity for its
ability to avoid the need for large-scale datasets. It requires the
sampling of positive and negative pairs from raw, unlabeled data.
During the learning process, it encourages convergence of the positive
pair representations in latent space while enforcing divergence of
the negative pairs. A prominent example is CLIP \cite{radford2021learning},
where it constructs positive pairs consisting of an image and a sentence
describing the same object, and negative pairs consist of an image
and a sentence that refer to different objects. While structurally
different, visual and text-based latent representations should contain
mutual information linked to the same concept. This strategy is also
applied to image-image pairs for data augmentation, e.g., SimCLR \cite{chen2020simple}.
The intuition is to bring different augmented views of the same image
closer in  latent space, while pushing different augmented views apart.
Proven highly effective for data representation, this technique has
pioneered subsequent works \cite{chen2021exploring,fan2024improving}
for robust feature representation learning. In this work, we applied
this technique to human activity recognition using multi-modal input,
enforcing convergence of latent representations from different modalities
originating from the same actor, despite their structural disparities.

%% file: method.tex
\subsection{Preliminaries \protect\label{subsec:Preliminaries}}

\textbf{Formulation:} Our goal is to design a model that leverages
multi-modal inputs, e.g., RGB sequences and human body key points,
for human activity recognition in videos. We achieve this by formulating
the problem under a \emph{neural query machine. }Our query machine
takes as input a human-centric query $\left\{ q_{i}\right\} $ that
probes different aspects regarding the movements of a specific human
actor and its relationships with the surrounding entities within a
video input $V$. The output is a prediction of an action label $\tilde{y}$,
based on the collective human-centric attributes in response to the
queries. Formally, our query machine is given as:

\begin{equation}
\tilde{y}=\mathcal{A}\left(\left\{ g^{m}\left(q_{i,t}^{m},V\right)\right\} _{m}\right).\label{eq:task_def}
\end{equation}
For each modality $m$, $q_{i,t}^{m}$ is $i$-th query at time step
$t$; $g^{m}\left(.\right)$ is a neural building block that retrieves
relevant information in $V$ in response to $q_{i,t}^{m};$ $\mathcal{A}\left(.\right)$
is a neural network that aggregates the attributes from the input
modalities and maps them to label space. 

Our work investigates human activity recognition in videos under two
specific applications: Spatio-Temporal Action Localization and Group
Activity Recognition. Since human activity is usually interpreted
through different layers of interactions, such as self movements and
cross-entity interactions, we hypothesize a \emph{compositional function}
for each modality-wise query machine $g^{m}\left(.\right)$. This
compositional design places humans at the center of relational modeling
of their interactions with the surroundings (Sec.~\ref{subsec:proposed_COMPUTER}). 

\textbf{Spatio-temporal video representation:} Following recent studies
\cite{wu2021towards,fan2021multiscale,yang2020temporal}, we first
extract a spatio-temporal representation for each video input $V$
using video feature extractors such as Slowfast \cite{feichtenhofer2019slowfast}
and MViT models \cite{fan2021multiscale,li2022mvitv2}. The video
$V$ is usually segmented into $T$ non-overlapping clips, resulting
in video features $X\in\mathbb{R}^{T\times FHW\times D}$, where $F$
is the number of frames in each clip, and $H,W,D$ are the height,
width, and channel dimension of the feature maps, respectively. 

\textbf{Query representation:} We use two input modalities as queries:
\emph{human-centric visual appearance} and \emph{body key points}. 

\emph{Human-centric visual appearance}: Visual appearance of human
actors themselves plays a crucial role in interpreting their actions.
To capture this, we follow \cite{sun2018actor,feichtenhofer2019slowfast}
to use an actor localization module to extract the appearance saliency
of human actors. First, for each video segment $t$ in the $T$ non-overlapping
clips from a video input, we utilize a human detector \cite{ren2015faster}
to localize human actors within their center frame. This yields a
set of bounding boxes for all $N$ detected actors. We then use RoI-Align
\cite{he2017mask} to extract visual appearance features for the $N$
actors: $Q_{t}^{\text{vis}}=\left\{ q_{i,t}^{\text{vis}}\mid q_{i,t}^{\text{vis}}\in\mathbb{R}^{1\times D}\right\} _{i=1}^{N}$.

\begin{figure}
\begin{centering}
\includegraphics[width=0.9\columnwidth]{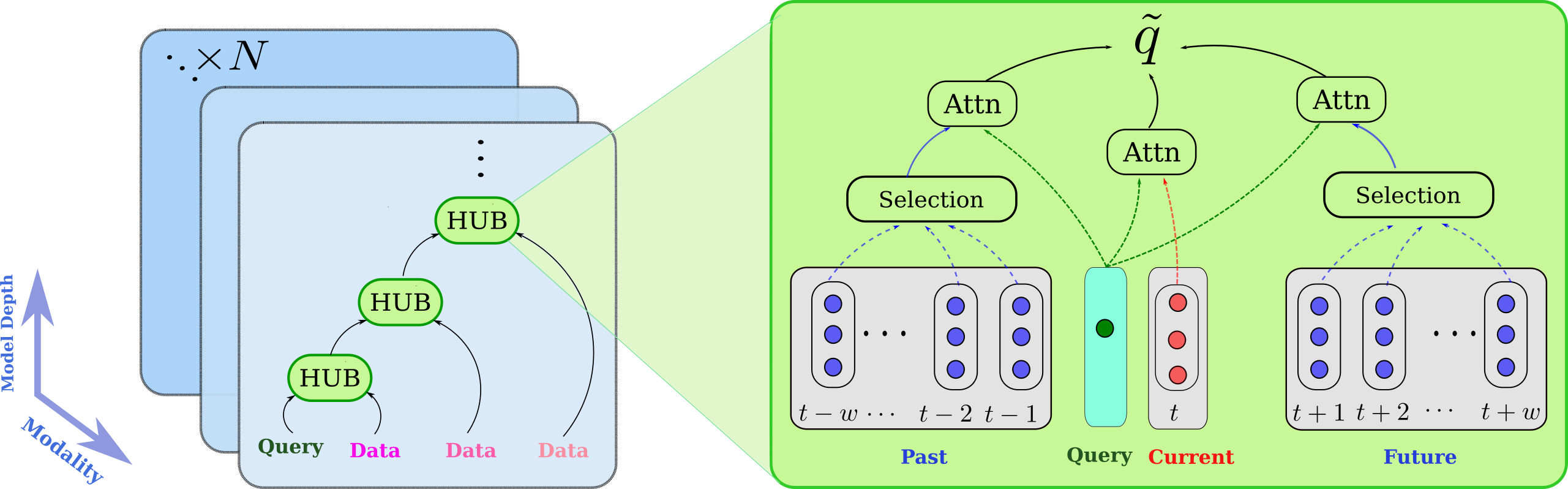}\vspace{4mm}
\par\end{centering}
\caption{$\protect\ModelName$ models the human-human and human-context interactions
in videos using a stack of $\protect\BlockName$ blocks. Each $\protect\BlockName$
takes as input a human-centric query $q_{i}$ (\textcolor{green}{green}
circle) of any input modalities and a knowledge base to iteratively
refine its knowledge about the human of interest. The knowledge base
is spatial-temporal features extracted from past (\textcolor{blue}{blue}
circles), current (\textcolor{red}{red} circles) and future (\textcolor{blue}{blue}
circles) video segments. Depending on the knowledge contained in the
knowledge base, whether it is human-centric features or general contextual
information, the $\protect\BlockName$ block can be used to flexibly
model the relationships between humans and their relationships with
the surrounding entities. Best viewed in color.\protect\label{fig:SURE}}
\end{figure}

\emph{Body key points}: We use the common framework Detectron2 \cite{wu2019detectron2}
to detect human body key points from RGB frames. Similar to visual
appearance feature extraction, we use the middle frame of a video
clip $t$ for pose detection, resulting in a set $Q_{t}^{\text{key}}$
of $N$ person skeletons: $Q_{t}^{\text{key}}=\left\{ q_{i,t}^{\text{key}}\mid q_{i,t}^{\text{key}}\in\mathbb{R}^{1\times D}\right\} _{i=1}^{N}$.

\subsection{Compositional Human-centric Query Machine\protect\label{subsec:proposed_COMPUTER}}

We propose a novel family of model architectures, dubbed \textbf{COMP}ositional
h\textbf{U}man-cen\textbf{T}ric qu\textbf{ER}y machine $(\ModelName)$,
for multi-modal human activity recognition in videos. $\ModelName$
leverages a modular design, combining several identical modality-wise
query machines that model the relationships between human actors and
their surroundings in both space and time. This modular design simplifies
the construction of $\ModelName$ by stacking identical building blocks,
facilitating dynamic model sizes and model's representation capabilities
for efficient action prediction.

Our query machines focus on two main types of interactions: \emph{human-human
interactions} and \emph{human-context interactions.} Inspired by \emph{Dang
et al.} \cite{dang2021hierarchical}, each modality-wise query machine
in $\ModelName$ adopts a two-stage design, where the output of the
first stage serves as the input for the second stage. Figure \ref{fig:SURE}
(on the left) provides a general architecture of $\ModelName$. One
of the key advantages of $\ModelName$'s modular design is its inherent
scalability. The system can be easily extended to incorporate additional
input modalities and handle different types of interactions. In this
work, we demonstrate this capability in the specific context of human
activity recognition with \emph{two modalities} (visual appearance
and body key points) and \emph{two types of interactions} (human-human
and human-context interactions). Mathematically, $\ModelName$ implements
each individual query machine $g^{m}\left(.\right)$ for the $m$-th
modality in Eq.~(\ref{eq:task_def}) using a compositional function:
\begin{equation}
g^{m}\left(q_{i,t}^{m},V\right)=\text{\ensuremath{\Phi_{c}}}\left(\Phi_{h}\left(q_{i,t}^{m},X^{H}\right),X^{C}\right).
\end{equation}
Here, $X^{H}$ and $X^{C}$ represent human-centric and contextual
features, respectively, derived from the embedding of the video input
$V$. We define $\Phi_{h}\left(.\right)$ and $\Phi_{c}\left(.\right)$
as reusable computational units called \textbf{HU}man-centric query
\textbf{B}locks ($\BlockName$s). These $\BlockName$ units play a
crucial role in modeling human-human interactions (HH-HUB)\textbf{
}and human-context interactions (HC-HUB). Since the operation of the
HUB is generic and does not depend on a specific modality, we omit
$m$ for the brevity. We elaborate the design of $\BlockName$ in
the following.

Central to the $\BlockName$'s operation is the widely used scaled
dot-product attention layer \cite{vaswani2017attention}:

\begin{equation}
\text{Attn}\left(q,K,V\right)=\sum_{\mu=1}^{M}\text{softmax}_{\mu}\left(\frac{K_{\mu}W_{k}\left(qW_{q}\right)}{\sqrt{d}}^{\top}\right)V_{\mu}W_{v},\label{eq:attention_layer}
\end{equation}
where query $q\in\mathbb{R}^{1\times D}$, keys $K\in\mathbb{R}^{M\times D}$,
values $V\in\mathbb{R}^{M\times D}$. The output of $\text{Attn}\left(q,K,V\right)$
is a vector in $\mathbb{R}^{1\times D}$ and $W_{q}\in\mathbb{R}^{D\times D}$,
$W_{k}\in\mathbb{R}^{D\times D}$, $W_{v}\in\mathbb{R}^{D\times D}$
are network parameters. Fig.~\ref{fig:SURE} (on the right) demonstrates
the operation of $\BlockName$. $\BlockName$ is comprised of stacked
attention layers that accounts for the similarity between the dynamics
of a human actor and its surroundings in space and time through \emph{three
information channels} (past, current, future). It searches for relevant
information of the query $q$ in memories $X_{\text{past}}$, $X_{\text{current}}$,
$X_{\text{future}}$ storing past, current and future knowledge in
the form of key-value pairs. While the query is a specific actor representation
at time $t$, the information in the memories can consist either human-centric
or general contextual information of a video clip at different points
in time. The output of $\BlockName$ is a refined representation of
the actor-specific feature in response to the given query. Denoting
$\tilde{q}_{i,t}$ as the output representation of actor $i$, defined
as:\vspace{-0.5cm}

\begin{align}
\tilde{q}{}_{i,t} & =\text{HUB}\left(q_{i,t},\left\{ X_{\text{past}},X_{\text{current}},X_{\text{future}}\right\} \right),
\end{align}
where $X_{\text{past}}=\left\{ K_{t-w:t-1},V_{t-w:t-1}\right\} $,
$X_{\text{\text{current}}}=\left\{ K_{t},V_{t}\right\} $, $X_{\text{future}}=\left\{ K_{t+1:t+w},V_{t+1:t+w}\right\} $
are key-value stores that encapsulates information in past, current
and future times. The window size $w$ indicates clips that are $w$
steps apart from the present clip $t$. To enable effective retrieval
of past/future information while reducing the computational costs,
we employ a \emph{pre-computed clip selection} mechanism that allows
us to skip irrelevant clips. In particular, we assess the relevance
of all clips within the window $w$ to the present clip at time $t$
using their feature similarity. We then select the top-$k$ most relevant
clips and store them as as key-value memories in the past $(X_{\text{past}}$)
and future times ($X_{\text{future}}$). $\BlockName$ computes each
pair $\left(q,X\right)$ using a multi-layer attention in Eq. (\ref{eq:attention_layer}),
followed by a linear aggregation layer which returns a single vector
$\tilde{q}{}_{i,t}$ for each human actor $i$.

\textbf{Human-human interactions with HH-HUB:} This stage considers
an actor in the relation with other actors involved in the same visual
scene in space and time. The HH-HUB $\Phi_{h}\left(.\right)$ takes
as input a query $q_{i,t}$, either visual appearance or human body
key points, representing an individual actor $i$ at video clip $t$
and three key-value stores $X_{\text{past}}^{\text{h}}$, $X_{\text{current}}^{\text{h}}$,
$X_{\text{future}}^{\text{h}}$ denoting visual appearance features
of all other human actors detected in past, present and future video
clips, respectively (See Sec.~\ref{subsec:Preliminaries}). While
the pair $\left(q_{i,t},X_{\text{current}}^{\text{h}}\right)$ at
current clip $t$ captures the spatial relationships between actors
in the current scene, the across-time pairs $\left(q_{i,t},X_{\text{past}}^{\text{h}}\right)$
and $\left(q_{i,t},X_{\text{future}}^{\text{h}}\right)$ provide information
about how the relationships evolve over time. The output of the HH-HUB
is a refined representation $\tilde{q}_{i,t}\in\mathbb{R}^{1\times D}$
for the actor $i$ at clip $t$: 
\begin{equation}
\tilde{q}_{i,t}=\left[\ensuremath{\mathcal{\text{Attn}}}\left(q_{i,t},X_{\text{past}}^{\text{h}}\right),\text{Attn}\left(q_{i,t},X_{\text{current}}^{\text{h}}\right),\text{Attn}\left(q_{i,t},X_{\text{future}}^{\text{h}}\right)\right]W_{a},\label{eq:hh-hub}
\end{equation}
where {[}· , ·{]} indicates feature concatenation, and $W_{a}\in\mathbb{R}^{3D\times D}$
is learnable parameters.

\textbf{Human-context interactions with HC-HUB:} Unlike the HH-HUB
module, the HC-HUB $\Phi_{c}\left(.\right)$ focuses on modeling human-context
relationships. It takes the output $\tilde{q}_{i,t}$ of the HH-HUB
as an input query and video spatio-temporal representations $X_{\text{past}}^{\text{c}}$,
$X_{\text{current}}^{\text{c}}$ and $X_{\text{future}}^{\text{c}}$
of past, present and future video clips (See Sec.~\ref{subsec:Preliminaries})
as key-value stores. The computation of the output $\hat{q}_{i,t}\in\mathbb{R}^{1\times D}$
of the HC-HUB is similar to the HH-HUB as in Eq. \ref{eq:hh-hub}.
It now incorporates both human-human and human-context interactions
over space and time.

\subsection{Cross-modality Consistency with Contrastive Loss}

\begin{figure}
\begin{centering}
\subfloat[$\protect\ModelName$ enhances baselines performance. \textcolor{blue}{Blue}
and \textcolor{lime}{green} bars are absolute point and percentage
improvement.\label{fig:backbone}]{\begin{centering}
\includegraphics[width=0.45\columnwidth]{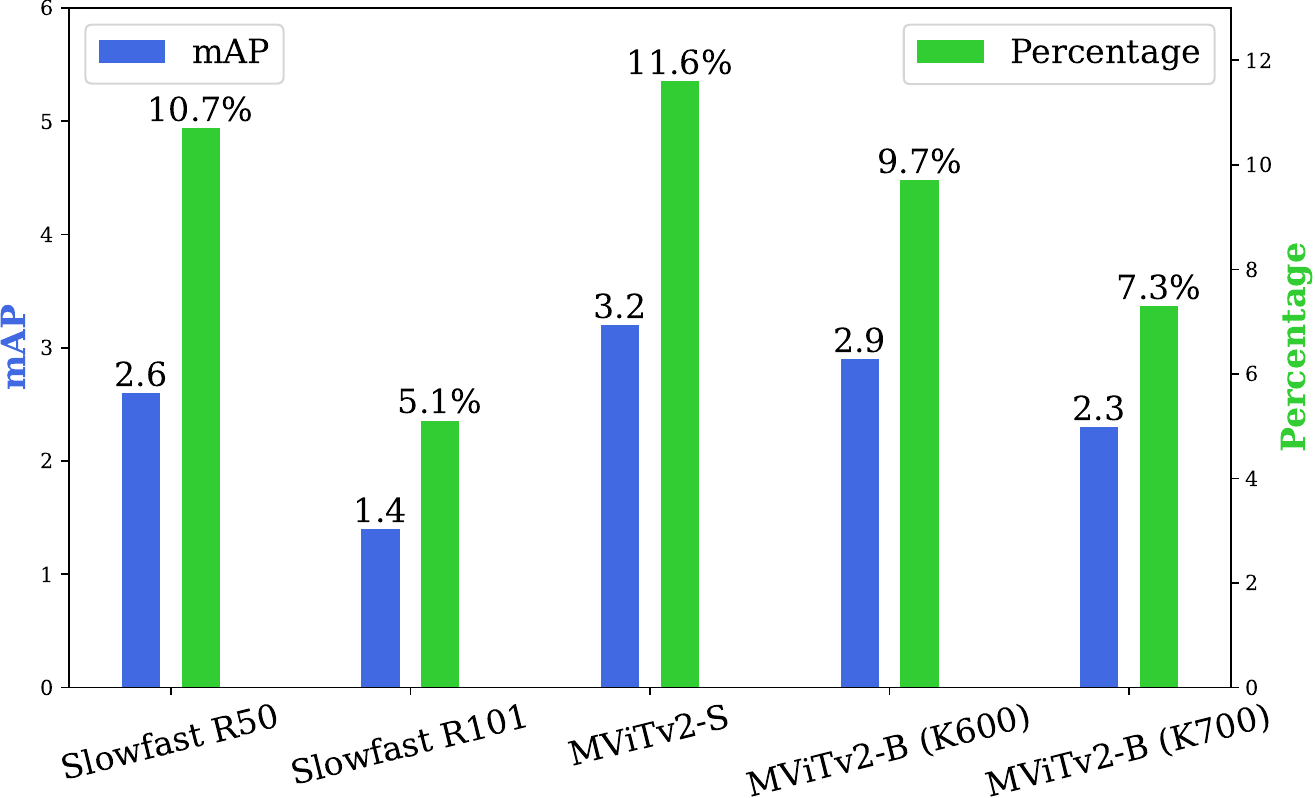}
\par\end{centering}
}\quad{}\subfloat[Combining multiple modalities improves model's capability for action
recognition.\label{fig:qualitative}]{\begin{centering}
\includegraphics[width=0.35\columnwidth]{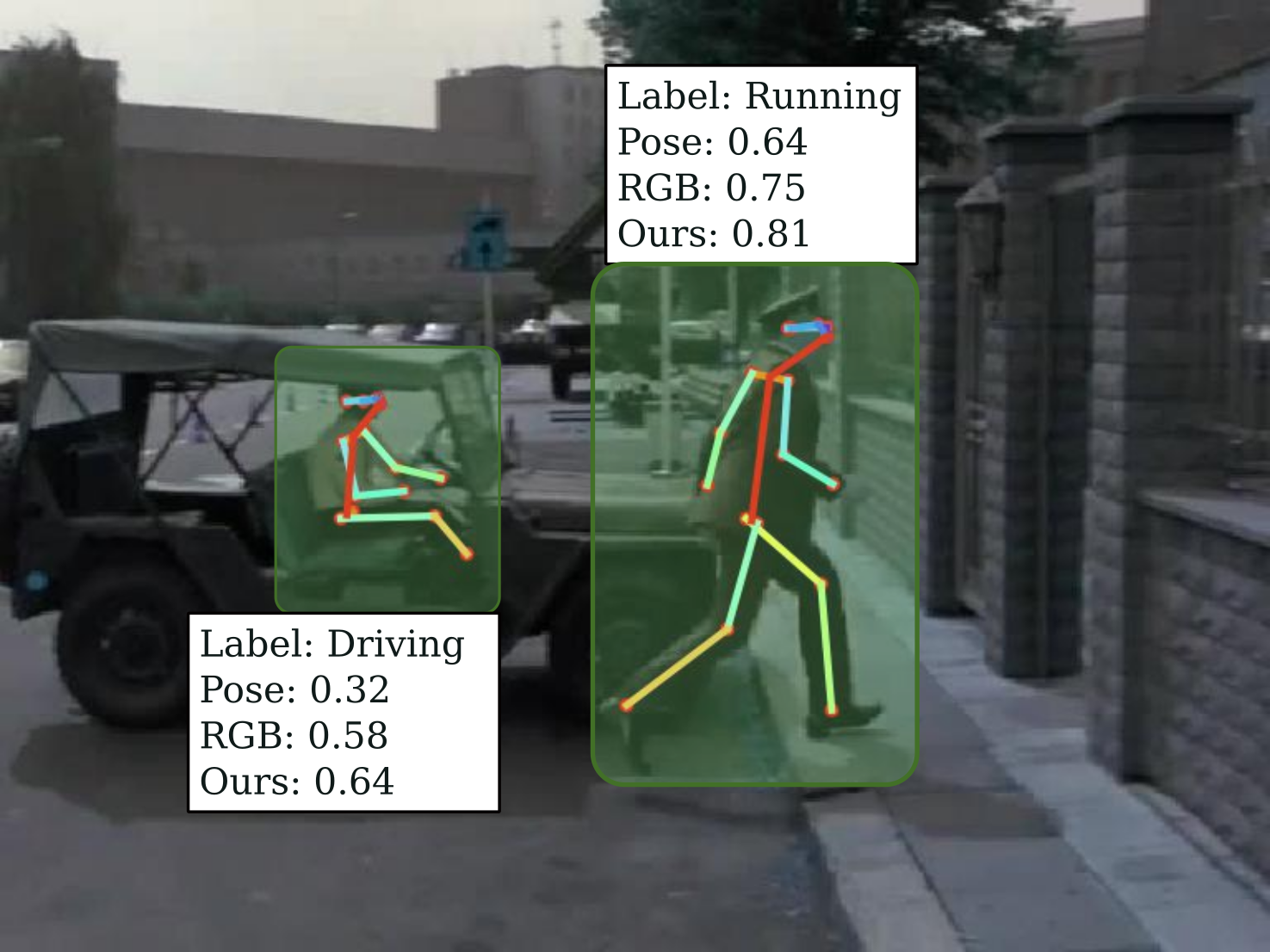}
\par\end{centering}
}
\par\end{centering}
\begin{centering}
\vspace{4mm}
\par\end{centering}
\caption{Quantitative and qualitative analysis of the proposed approach on
the AVA dataset.}
\end{figure}

In multi-modal human activity recognition, models should leverage
complementary aspects across modalities for label prediction. However,
inherent representation disparities between input modalities make
finding a joint representation capturing the saliency across all modalities
challenging. Instead of directly fusing high-level features of these
modalities together, we introduce a consistency loss to encourage
the model to exploit mutual information across input modalities of
the same person, as they both lead to the same activity prediction.
We achieve this by maximizing the mutual information between any pairs
of input modalities. Specifically, we sample a positive pair by taking
the final representations $\hat{q}_{i,t}^{\text{vis}},\hat{q}_{i,t}^{\text{key}}$
by $\ModelName$, which belongs to the same person while treating
$k$ augmented samples randomly paired from different individuals
within a mini-batch as negative samples. Our cross-modality consistency
loss $\mathcal{L_{\text{CC}}}$ is implemented similar to the contrastive
loss in \cite{chen2020simple}:
\begin{equation}
\ensuremath{\mathcal{L}_{\text{CC}}=-\log\frac{\exp\left(\text{sim}\left(\hat{q}_{i,t}^{\text{vis}},\hat{q}_{i,t}^{\text{key}}\right)\right)}{\sum_{k=1}^{B}\mathbb{\mathds{I}}{}_{[k\neq i]}\exp\left(\text{sim}\left(\hat{q}_{i,t}^{\text{vis}},\hat{q}_{k,t}^{\text{key}}\right)\right)}},
\end{equation}
where, $\text{sim}\left(\cdot,\cdot\right)$ is the cosine similarity
function between two input vectors. $\mathbb{\mathds{I}}(.)$ is an
indicator function iff $k\neq i$ within mini batch $B$. Importantly,
our consistency loss allows us to train the proposed model in an unsupervised
manner without the need for additional training data. We train our
models with this consistency loss together with the usual cross-entropy
loss for label prediction. We detail the training of our two tasks
as below.

\textbf{Spatio-temporal action localization:} We use a classifier
of an MLP followed by a logistic function to predict action labels
by an actor at time step. Our network is trained end-to-end by jointly
minimizing the binary cross entropy loss and the consistency loss:
$\mathcal{L}=\mathcal{L_{\text{BCE}}}+\mathcal{L}_{\text{CC}}$. 

\textbf{Group activity recognition:} As all actors share the same
action label throughout a video input, we first apply the arithmetic
mean function across actors and along the temporal axis on the actor-specific
representations $\hat{q}^{\text{vis}}$ and $\hat{q}^{\text{key}}$
to obtain a single output vector. We then use an MLP layer to map
the video feature to label space before applying the soft-max function
to return action label probabilities.We jointly minimize the cross
entropy loss and the consistency loss to train the network.

%% file: exp.tex
We evaluate the effectiveness of the proposed framework on two major
applications of human behavior understanding in videos: Spatio-Temporal
Action Localization on AVA v2.2 \cite{gu2018ava} and Group Activity
Recognition on Collective Activity dataset \cite{Choi2009WhatAT}.

\subsection{Spatio-temporal Action Localization \protect\label{subsec:Action-Detection}}

\textbf{}

\textbf{Quantitative results}: We first demonstrate the efficacy of
$\ModelName$ when using video features by different video representation
backbones. The results are displayed in Fig.~\ref{fig:backbone}.
In general, $\ModelName$ consistently improves all the baselines
where the gaps are more significant on weaker baselines. 

\begin{table}
\begin{centering}
\begin{tabular}{>{\centering}p{0.1\textwidth}|>{\centering}p{0.22\textwidth}|>{\centering}p{0.06\textwidth}||>{\raggedright}m{0.1\textwidth}|>{\centering}p{0.22\textwidth}|>{\centering}p{0.06\textwidth}}
\hline 
Pretrained & Method & mAP & Pretrained & Method & mAP\tabularnewline
\hline 
\hline 
\multirow{8}{0.1\textwidth}{K400} & Slowfast R50 \cite{feichtenhofer2019slowfast} & 22.7 & \multirow{4}{0.1\textwidth}{K600} & OT \cite{wu2021towards} & 31.0\tabularnewline
 & SlowFast R101 \cite{feichtenhofer2019slowfast} & 23.8 &  & ACAR R101\cite{pan2021actor} & 31.8\tabularnewline
 & ORViT \cite{Herzig2022Orvit} & 26.6 &  & MViTv2-B \cite{li2022mvitv2} & 30.5\tabularnewline
 & MemViT \cite{wu2022memvit} & 29.3 &  & $\boldsymbol{\ModelName}$ & \textbf{32.6}\tabularnewline
\cline{4-6}
 & MViTv1-B \cite{li2022mvitv2} & 27.3 & \multirow{4}{0.1\textwidth}{K700} & AIA\cite{tang2020asynchronous} & 32.3\tabularnewline
 & MViTv2-S \cite{li2022mvitv2} & 27.6 &  & HIT \cite{faure2023holistic} & 32.6\tabularnewline
 & MViTv2-B \cite{li2022mvitv2} & 29.0 &  & MViTv2-B \cite{li2022mvitv2} & 31.3\tabularnewline
 & $\boldsymbol{\ModelName}$ & \textbf{30.8} &  & $\boldsymbol{\ModelName}$ & \textbf{33.6}\tabularnewline
\hline 
\end{tabular}
\par\end{centering}
\begin{centering}
\vspace{4mm}
\par\end{centering}
\caption{Comparison against the state-of-the-art methods on AVA dataset. \protect\label{tab:with-otherAVA}}
\end{table}

We also compare $\ModelName$ against the most recent SoTA methods
on AVA (See Tab. \ref{tab:with-otherAVA}). We categorize the prior
works based on the respective datasets that their video feature extractors
are pre-trained on, following \cite{wu2022memvit,faure2023holistic}.
As seen, $\ModelName$ consistently outperforms all the recent approaches
across all categories. While these works either only focus on human-human
interactions such as \cite{wu2021towards} or human-object/human-context
interactions as in \cite{tang2020asynchronous,pan2021actor}, $\ModelName$
enjoys the benefits of these two types of interactions within a single
model. While $\ModelName$ clearly outperform approaches using single
modalities such as MViTv2 \cite{li2022mvitv2}, ORViT \cite{Herzig2022Orvit}
and MemViT \cite{wu2022memvit}, we wish to emphasize its superior
performance when comparing with the most recent approach HIT \cite{faure2023holistic}
that leverages identical input modalities. This clearly demonstrates
the effectiveness of our proposed method in both architecture modeling
with $\BlockName$ units and learning with the cross-modality consistency
loss.

\textbf{Qualitative results:} We showcase examples taken from the
AVA dataset in Fig.~\ref{fig:qualitative}. Combining both modalities
significantly enhances prediction performance compared to using a
single modality. Actions that requires efficient temporal modeling
such as running and driving are among the ones that benefit the most
from leveraging human body key points. 

\begin{table}
\begin{centering}
\begin{tabular}{>{\centering}p{0.042\columnwidth}|>{\centering}p{0.25\columnwidth}|>{\centering}p{0.08\columnwidth}|c|c|c}
\hline 
No. & Method & mAP & FLOP (G) & Infer. time (s) & Params (M)\tabularnewline
\hline 
\hline 
1 & MViTv2-S & 27.6 & 64.5 & 1.27 & 34.3\tabularnewline
1.a & MViTv2-S + 6 attns & 28.3 & 65.0 & 1.29 & 48.6\tabularnewline
1.b & MViTv2-S + 12 attns & 28.2 & 66.1 & 1.30 & 62.6\tabularnewline
1.c & $\boldsymbol{\ModelName}$ & 30.8 & 68.4 & 1.31 & 54.3\tabularnewline
\hline 
2 & MViTv2-B & 29.0 & 225.2 & 1.62 & 51.0\tabularnewline
2.a & MViTv2-S + 6 attns & 29.8 & 226.1 & 1.65 & 65.0\tabularnewline
2.b & MViTv2-S + 12 attns & 29.9 & 228.0 & 1.66 & 79.3\tabularnewline
2.c & $\boldsymbol{\ModelName}$ & 32.6 & 230.0 & 1.66 & 70.9\tabularnewline
\hline 
\end{tabular}
\par\end{centering}
\centering{}\vspace{4mm}
\caption{Trade-off between performance (mAP) and computation cost (in FLOPs,
Inference time and No. of parameters) when comparing $\protect\ModelName$
with different baselines. \protect\label{tab:computational}}
\end{table}

\textbf{Ablation studies}: We conduct a comprehensive analysis on
$\ModelName$'s computational costs (Tab.~\ref{tab:computational})
and the contributions of each input modality (Tab.~\ref{tab:modality}).
We also provide additional analysis on the effects of ablating different
designated components from the full design in the Supp. All ablation
studies use the MViTv2-S backbone.

\textbf{\emph{Computational complexity:}} To demonstrate the benefits
of $\ModelName$, we compare it with stronger baselines of similar
representation capacity (a.k.a model size) in Tab.~\ref{tab:computational}.
These baselines are implemented by fine-tuning extending MViTv2 baselines
with additional self-attention layers on AVA v2. Results show that
simply increasing model size offer minimal improvement (See Row 1
vs. 1.a/1.b, and Row 2 vs. 2.a/2.b). In contrast, COMPUTER significantly
enhances baseline performance with minimal additional costs. Specifically,
$\ModelName$ improves MViTv2-S by 3.0 points (10.9\%) , with only
6.0\% increase in GFLOPs and around 3.0\% additional inference time.
We obverse consistent behaviors on MViTv2-B. Importantly, $\ModelName$
with MViTv2-S baseline even outperforms MViTv2-B despite faster inference
time and only 1/3 of the GFLOPs, thanks to the sparsity of our human
input tokens (Row 1.c vs Row 2).

\textbf{\emph{Effectiveness of each modality:}} We analyze the impact
of each modality on the performance in Tab.~\ref{tab:modality}.
RGB sequences slightly outperform body key points thanks to its rich
information. $\ModelName$ successfully leverages the advantages of
each modality to improve the performance when using them in combination
(Row 3). Additionally, the proposed consistency loss considerably
improves the performance by nearly 1.5 points (5.1\%). 

\begin{figure*}

\begin{minipage}[t]{0.49\columnwidth}%
\begin{table}[H]
\begin{centering}
\begin{tabular}{>{\centering}p{1cm}>{\centering}p{1cm}>{\centering}p{1.6cm}|c}
\hline 
Skeleton & RGB & Consistency loss & mAP\tabularnewline
\hline 
\hline 
\ding{52} &  &  & 28.2\tabularnewline
 & \ding{52} &  & 28.9\tabularnewline
\ding{52} & \ding{52} &  & 29.4\tabularnewline
\hline 
\ding{52} & \ding{52} & \ding{52} & \textbf{30.8}\tabularnewline
\hline 
\end{tabular}
\par\end{centering}
\begin{centering}
\vspace{4mm}
\par\end{centering}
\caption{Ablation on the effectiveness of each modality\protect\label{tab:modality}}
\end{table}
\end{minipage}\hfill{}%
\begin{minipage}[t]{0.49\columnwidth}%
\begin{table}[H]
\begin{centering}
\begin{tabular}{>{\centering}p{3.8cm}|>{\centering}p{1.5cm}}
\hline 
Method & Test Acc.\tabularnewline
\hline 
\hline 
Baseline (InceptionNet) & 86.0\%\tabularnewline
CERN \cite{shu2017cern} & 87.2\%\tabularnewline
SBGAR \cite{SBGAR} & 86.1\%\tabularnewline
GT \cite{wu2019learning} & 91.0\%\tabularnewline
\hline 
\textbf{Ours} & \textbf{93.3\%}\tabularnewline
\hline 
\end{tabular}
\par\end{centering}
\begin{centering}
\vspace{4mm}
\par\end{centering}
\caption{Comparison against the state-of-the-art methods on Collective Activity
dataset.\protect\label{tab:other-GAR}}
\end{table}
\end{minipage}

\end{figure*} 

\begin{table}
\begin{centering}
\begin{tabular}{>{\centering}p{5.7cm}|>{\centering}p{1.5cm}}
\hline 
Method & mAP\tabularnewline
\hline 
\hline 
MViTv2-S baseline & 27.6\tabularnewline
W/o hierarchy design & 29.3\tabularnewline
W/o HC-HUB & 28.7\tabularnewline
W/o HH-HUB & 28.9\tabularnewline
W/o temporal modeling & 29.2\tabularnewline
W/o pre-computed clip selection & 29.6\tabularnewline
\hline 
Full $\ModelName$ & \textbf{30.8}\tabularnewline
\hline 
\end{tabular}\vspace{4mm}
\par\end{centering}
\caption{Ablation on the effectiveness of $\protect\ModelName$'s components\protect\label{tab:component-SURE}}
\end{table}

\textbf{\emph{Effectiveness of each component in $\ModelName$:}}

To provide more insights of our architecture $\ModelName$, we ablate
its components and observe its effect to the overall performance.
In general, ablating any designed components of $\ModelName$ would
result in degradation in performance (See Table \ref{tab:component-SURE}). 

\emph{\quad{}Effectiveness of the hierarchy design: }In this experiment,
the target human query attends to both the human and context elements
simultaneously, without imposing a hierarchical order. The significant
performance drop by 1.5 points (5.1\%) highlights the importance of
our hierarchical design.

\emph{\quad{}Effectiveness of the HC-HUB block: }This experiment
removes all HC-HUB blocks out of the original design of $\ModelName$.
This leads to a considerable decrease in performance by nearly 2.0
points (5.9\%).

\emph{\quad{}Effectiveness of the HH-HUB block:} Similarly, this
experiment removes all the HH-HUB blocks. With the absence of the
human-human interactions, we observe a similar level of performance
degradation.

\emph{\quad{}Effectiveness of temporal modeling:} This experiment
limits all $\BlockName$ blocks to consider only the present information
channel while ignoring the other channels in past and future times.
Without considering the temporal dynamics of information, the performance
drops by nearly 1.5 points (4.8\%).

\emph{\quad{}Effectiveness of pre-computed clip selection:} This
experiment justifies the benefit of our pre-computed clip selection.
Instead of selectively choosing top$-k$ top relevant past/future
clips with the clip at present time, we take into account all video
clips within the window size $w$ and merely take average over the
post attention layer outputs. This suffers from 1.0 points performance
decrease that highlights the necessity of the information selection
strategy for the sake of both performance and computational burden.

\subsection{Group Activity Recognition}

The Collective Activity dataset \cite{Choi2009WhatAT} includes 44
clips of five types of group activities including crossing, queuing,
walking, waiting and talking. For fair comparisons with prior works
\cite{wu2019learning,hsu2017unsupervised,SBGAR}, we use InceptionNet
\cite{inception} pre-trained on ImageNet \cite{ILSVRC15} for feature
extraction.

\textbf{Quantitative results:} The results of our proposed $\ModelName$
model for action group recognition, shown in Tab.~\ref{tab:other-GAR},
demonstrate the effectiveness of our approach. Our method clearly
outperforms existing works by successfully incorporating multiple
modalities, leading to a more comprehensive representation.

%% file: conclusion.tex
We introduced a unified framework named $\ModelName$ for multi-modal
human activity recognition. The framework features a generic architecture
effectively retrieving information about human movements and the relationships
between human actors and their surroundings from different input modalities.
We also introduced a novel consistency loss to leverage the complementary
information across modalities for robust prediction of human activity
in an unsupervised manner. Through extensive experiments on two applications,
our framework demonstrated high efficiency approach compared to existing
methods.

%% file: main.bbl
\begin{thebibliography}{33}
\providecommand{\natexlab}[1]{#1}
\providecommand{\url}[1]{\texttt{#1}}
\expandafter\ifx\csname urlstyle\endcsname\relax
  \providecommand{\doi}[1]{doi: #1}\else
  \providecommand{\doi}{doi: \begingroup \urlstyle{rm}\Url}\fi

\bibitem[Carreira and Zisserman(2017)]{carreira2017quo}
Joao Carreira and Andrew Zisserman.
\newblock Quo vadis, action recognition? a new model and the kinetics dataset.
\newblock In \emph{proceedings of the IEEE Conference on Computer Vision and Pattern Recognition}, pages 6299--6308, 2017.

\bibitem[Chen et~al.(2020)Chen, Kornblith, Norouzi, and Hinton]{chen2020simple}
Ting Chen, Simon Kornblith, Mohammad Norouzi, and Geoffrey Hinton.
\newblock A simple framework for contrastive learning of visual representations.
\newblock In \emph{International conference on machine learning}, pages 1597--1607. PMLR, 2020.

\bibitem[Chen and He(2021)]{chen2021exploring}
Xinlei Chen and Kaiming He.
\newblock Exploring simple siamese representation learning.
\newblock In \emph{Proceedings of the IEEE/CVF conference on computer vision and pattern recognition}, pages 15750--15758, 2021.

\bibitem[Choi et~al.(2009)Choi, Shahid, and Savarese]{Choi2009WhatAT}
Wongun Choi, Khuram Shahid, and Silvio Savarese.
\newblock What are they doing? : Collective activity classification using spatio-temporal relationship among people.
\newblock \emph{2009 IEEE 12th International Conference on Computer Vision Workshops, ICCV Workshops}, pages 1282--1289, 2009.

\bibitem[Dang et~al.(2021)Dang, Le, Le, and Tran]{dang2021hierarchical}
Long~Hoang Dang, Thao~Minh Le, Vuong Le, and Truyen Tran.
\newblock Hierarchical object-oriented spatio-temporal reasoning for video question answering.
\newblock \emph{IJCAI}, 2021.

\bibitem[Duan et~al.(2022)Duan, Zhao, Chen, Lin, and Dai]{duan2022revisiting}
Haodong Duan, Yue Zhao, Kai Chen, Dahua Lin, and Bo~Dai.
\newblock Revisiting skeleton-based action recognition.
\newblock In \emph{Proceedings of the IEEE/CVF conference on computer vision and pattern recognition}, pages 2969--2978, 2022.

\bibitem[Fan et~al.(2021)Fan, Xiong, Mangalam, Li, Yan, Malik, and Feichtenhofer]{fan2021multiscale}
Haoqi Fan, Bo~Xiong, Karttikeya Mangalam, Yanghao Li, Zhicheng Yan, Jitendra Malik, and Christoph Feichtenhofer.
\newblock Multiscale vision transformers.
\newblock In \emph{ICCV}, 2021.

\bibitem[Fan et~al.(2024)Fan, Krishnan, Isola, Katabi, and Tian]{fan2024improving}
Lijie Fan, Dilip Krishnan, Phillip Isola, Dina Katabi, and Yonglong Tian.
\newblock Improving clip training with language rewrites.
\newblock \emph{Advances in Neural Information Processing Systems}, 36, 2024.

\bibitem[Faure et~al.(2023)Faure, Chen, and Lai]{faure2023holistic}
Gueter~Josmy Faure, Min-Hung Chen, and Shang-Hong Lai.
\newblock Holistic interaction transformer network for action detection.
\newblock In \emph{Proceedings of the IEEE/CVF Winter Conference on Applications of Computer Vision}, pages 3340--3350, 2023.

\bibitem[Feichtenhofer et~al.(2019)Feichtenhofer, Fan, Malik, and He]{feichtenhofer2019slowfast}
Christoph Feichtenhofer, Haoqi Fan, Jitendra Malik, and Kaiming He.
\newblock Slowfast networks for video recognition.
\newblock In \emph{Proceedings of the IEEE/CVF international conference on computer vision}, pages 6202--6211, 2019.

\bibitem[Gu et~al.(2018)Gu, Sun, Ross, Vondrick, Pantofaru, Li, Vijayanarasimhan, Toderici, Ricco, Sukthankar, et~al.]{gu2018ava}
Chunhui Gu, Chen Sun, David~A Ross, Carl Vondrick, Caroline Pantofaru, Yeqing Li, Sudheendra Vijayanarasimhan, George Toderici, Susanna Ricco, Rahul Sukthankar, et~al.
\newblock Ava: A video dataset of spatio-temporally localized atomic visual actions.
\newblock In \emph{Proceedings of the IEEE conference on computer vision and pattern recognition}, pages 6047--6056, 2018.

\bibitem[He et~al.(2017)He, Gkioxari, Doll{\'a}r, and Girshick]{he2017mask}
Kaiming He, Georgia Gkioxari, Piotr Doll{\'a}r, and Ross Girshick.
\newblock Mask r-cnn.
\newblock In \emph{Proceedings of the IEEE international conference on computer vision}, pages 2961--2969, 2017.

\bibitem[Herzig et~al.(2022)Herzig, Ben-Avraham, Mangalam, Bar, Chechik, Rohrbach, Darrell, and Globerson]{Herzig2022Orvit}
Roei Herzig, Elad Ben-Avraham, Karttikeya Mangalam, Amir Bar, Gal Chechik, Anna Rohrbach, Trevor Darrell, and Amir Globerson.
\newblock Object-region video transformers.
\newblock In \emph{Proceedings of the IEEE/CVF Conference on Computer Vision and Pattern Recognition (CVPR)}, pages 3148--3159, June 2022.

\bibitem[Hsu et~al.(2017)Hsu, Zhang, and Glass]{hsu2017unsupervised}
Wei-Ning Hsu, Yu~Zhang, and James Glass.
\newblock Unsupervised learning of disentangled and interpretable representations from sequential data.
\newblock In \emph{Advances in neural information processing systems}, pages 1878--1889, 2017.

\bibitem[Li and Chuah(2017)]{SBGAR}
Xin Li and Mooi~Choo Chuah.
\newblock Sbgar: Semantics based group activity recognition.
\newblock In \emph{2017 IEEE International Conference on Computer Vision (ICCV)}, pages 2895--2904, 2017.
\newblock \doi{10.1109/ICCV.2017.313}.

\bibitem[Li et~al.(2022)Li, Wu, Fan, Mangalam, Xiong, Malik, and Feichtenhofer]{li2022mvitv2}
Yanghao Li, Chao-Yuan Wu, Haoqi Fan, Karttikeya Mangalam, Bo~Xiong, Jitendra Malik, and Christoph Feichtenhofer.
\newblock Mvitv2: Improved multiscale vision transformers for classification and detection.
\newblock In \emph{Proceedings of the IEEE/CVF Conference on Computer Vision and Pattern Recognition}, pages 4804--4814, 2022.

\bibitem[Pan et~al.(2021)Pan, Chen, Shou, Liu, Shao, and Li]{pan2021actor}
Junting Pan, Siyu Chen, Mike~Zheng Shou, Yu~Liu, Jing Shao, and Hongsheng Li.
\newblock Actor-context-actor relation network for spatio-temporal action localization.
\newblock In \emph{Proceedings of the IEEE/CVF Conference on Computer Vision and Pattern Recognition}, pages 464--474, 2021.

\bibitem[Radford et~al.(2021)Radford, Kim, Hallacy, Ramesh, Goh, Agarwal, Sastry, Askell, Mishkin, Clark, et~al.]{radford2021learning}
Alec Radford, Jong~Wook Kim, Chris Hallacy, Aditya Ramesh, Gabriel Goh, Sandhini Agarwal, Girish Sastry, Amanda Askell, Pamela Mishkin, Jack Clark, et~al.
\newblock Learning transferable visual models from natural language supervision.
\newblock In \emph{International conference on machine learning}, pages 8748--8763. PMLR, 2021.

\bibitem[Ren et~al.(2015)Ren, He, Girshick, and Sun]{ren2015faster}
Shaoqing Ren, Kaiming He, Ross Girshick, and Jian Sun.
\newblock Faster r-cnn: Towards real-time object detection with region proposal networks.
\newblock In \emph{NIPS}, pages 91--99, 2015.

\bibitem[Russakovsky et~al.(2015)Russakovsky, Deng, Su, Krause, Satheesh, Ma, Huang, Karpathy, Khosla, Bernstein, Berg, and Fei-Fei]{ILSVRC15}
Olga Russakovsky, Jia Deng, Hao Su, Jonathan Krause, Sanjeev Satheesh, Sean Ma, Zhiheng Huang, Andrej Karpathy, Aditya Khosla, Michael Bernstein, Alexander~C. Berg, and Li~Fei-Fei.
\newblock {ImageNet Large Scale Visual Recognition Challenge}.
\newblock \emph{International Journal of Computer Vision (IJCV)}, 115\penalty0 (3):\penalty0 211--252, 2015.
\newblock \doi{10.1007/s11263-015-0816-y}.

\bibitem[Shi et~al.(2019)Shi, Zhang, Cheng, and Lu]{shi2019skeleton}
Lei Shi, Yifan Zhang, Jian Cheng, and Hanqing Lu.
\newblock Skeleton-based action recognition with directed graph neural networks.
\newblock In \emph{Proceedings of the IEEE/CVF conference on computer vision and pattern recognition}, pages 7912--7921, 2019.

\bibitem[Shu et~al.(2017)Shu, Todorovic, and Zhu]{shu2017cern}
Tianmin Shu, Sinisa Todorovic, and Song-Chun Zhu.
\newblock Cern: confidence-energy recurrent network for group activity recognition.
\newblock In \emph{Proceedings of the IEEE conference on computer vision and pattern recognition}, pages 5523--5531, 2017.

\bibitem[Su et~al.(2019)Su, Ouyang, Zhou, and Xu]{Su2019Improving}
Rui Su, Wanli Ouyang, Luping Zhou, and Dong Xu.
\newblock Improving action localization by progressive cross-stream cooperation.
\newblock In \emph{Proceedings of the IEEE/CVF Conference on Computer Vision and Pattern Recognition}, pages 12016--12025, 2019.

\bibitem[Sun et~al.(2018)Sun, Shrivastava, Vondrick, Murphy, Sukthankar, and Schmid]{sun2018actor}
Chen Sun, Abhinav Shrivastava, Carl Vondrick, Kevin Murphy, Rahul Sukthankar, and Cordelia Schmid.
\newblock Actor-centric relation network.
\newblock In \emph{Proceedings of the European Conference on Computer Vision (ECCV)}, pages 318--334, 2018.

\bibitem[Szegedy et~al.(2015)Szegedy, Liu, Jia, Sermanet, Reed, Anguelov, Erhan, Vanhoucke, and Rabinovich]{inception}
Christian Szegedy, Wei Liu, Yangqing Jia, Pierre Sermanet, Scott Reed, Dragomir Anguelov, Dumitru Erhan, Vincent Vanhoucke, and Andrew Rabinovich.
\newblock Going deeper with convolutions.
\newblock In \emph{2015 IEEE Conference on Computer Vision and Pattern Recognition (CVPR)}, pages 1--9, 2015.
\newblock \doi{10.1109/CVPR.2015.7298594}.

\bibitem[Tang et~al.(2020)Tang, Xia, Mu, Pang, and Lu]{tang2020asynchronous}
Jiajun Tang, Jin Xia, Xinzhi Mu, Bo~Pang, and Cewu Lu.
\newblock Asynchronous interaction aggregation for action detection.
\newblock \emph{arXiv preprint arXiv:2004.07485}, 2020.

\bibitem[Vaswani et~al.(2017)Vaswani, Shazeer, Parmar, Uszkoreit, Jones, Gomez, Kaiser, and Polosukhin]{vaswani2017attention}
Ashish Vaswani, Noam Shazeer, Niki Parmar, Jakob Uszkoreit, Llion Jones, Aidan~N Gomez, {\L}ukasz Kaiser, and Illia Polosukhin.
\newblock Attention is all you need.
\newblock \emph{Advances in neural information processing systems}, 30, 2017.

\bibitem[Wu and Krahenbuhl(2021)]{wu2021towards}
Chao-Yuan Wu and Philipp Krahenbuhl.
\newblock Towards long-form video understanding.
\newblock In \emph{Proceedings of the IEEE/CVF Conference on Computer Vision and Pattern Recognition}, pages 1884--1894, 2021.

\bibitem[Wu et~al.(2022)Wu, Li, Mangalam, Fan, Xiong, Malik, and Feichtenhofer]{wu2022memvit}
Chao-Yuan Wu, Yanghao Li, Karttikeya Mangalam, Haoqi Fan, Bo~Xiong, Jitendra Malik, and Christoph Feichtenhofer.
\newblock Memvit: Memory-augmented multiscale vision transformer for efficient long-term video recognition.
\newblock In \emph{Proceedings of the IEEE/CVF Conference on Computer Vision and Pattern Recognition}, pages 13587--13597, 2022.

\bibitem[Wu et~al.(2019{\natexlab{a}})Wu, Wang, Wang, Guo, and Wu]{wu2019learning}
Jianchao Wu, Limin Wang, Li~Wang, Jie Guo, and Gangshan Wu.
\newblock Learning actor relation graphs for group activity recognition.
\newblock In \emph{Proceedings of the IEEE/CVF Conference on computer vision and pattern recognition}, pages 9964--9974, 2019{\natexlab{a}}.

\bibitem[Wu et~al.(2020)Wu, Kuang, Wang, Zhang, and Wu]{wu2020context}
Jianchao Wu, Zhanghui Kuang, Limin Wang, Wayne Zhang, and Gangshan Wu.
\newblock Context-aware rcnn: A baseline for action detection in videos.
\newblock In \emph{Computer Vision--ECCV 2020: 16th European Conference, Glasgow, UK, August 23--28, 2020, Proceedings, Part XXV 16}, pages 440--456. Springer, 2020.

\bibitem[Wu et~al.(2019{\natexlab{b}})Wu, Kirillov, Massa, Lo, and Girshick]{wu2019detectron2}
Yuxin Wu, Alexander Kirillov, Francisco Massa, Wan-Yen Lo, and Ross Girshick.
\newblock Detectron2.
\newblock \url{https://github.com/facebookresearch/detectron2}, 2019{\natexlab{b}}.

\bibitem[Yang et~al.(2020)Yang, Xu, Shi, Dai, and Zhou]{yang2020temporal}
Ceyuan Yang, Yinghao Xu, Jianping Shi, Bo~Dai, and Bolei Zhou.
\newblock Temporal pyramid network for action recognition.
\newblock In \emph{Proceedings of the IEEE/CVF conference on computer vision and pattern recognition}, pages 591--600, 2020.

\end{thebibliography}
